\documentclass{article}

\usepackage{microtype}
\usepackage{graphicx}
\usepackage{subcaption}
\usepackage{booktabs}
\usepackage[utf8]{inputenc}
\usepackage[T1]{fontenc}

\usepackage{arxiv}
\setcitestyle{authoryear,open={(},close={)}}
\usepackage[subtle]{savetrees}
\usepackage{titling}

\usepackage{xcolor}
\usepackage{fontawesome5}
\usepackage{inconsolata}
\usepackage{capt-of}
\usepackage{colortbl}
\usepackage{multirow}
\usepackage{array}
\usepackage{tabularx}
\usepackage{amsmath}
\usepackage{amssymb}
\usepackage{float}

\definecolor{humanlyrose}{HTML}{B08D84}
\definecolor{humanlyrosedark}{HTML}{725B55}
\definecolor{humanlyroselight}{HTML}{F3EBE8}
\definecolor{humanlyrow}{HTML}{F7F1EF}
\definecolor{surveyusability}{HTML}{7F977F}
\definecolor{surveyrole}{HTML}{A28C55}

\newcommand{\yesfeature}{\ensuremath{\checkmark}}
\newcommand{\nofeature}{}
\newcolumntype{Y}{>{\raggedright\arraybackslash}X}
\newcolumntype{M}[1]{>{\raggedright\arraybackslash}m{#1}}
\newcolumntype{N}[1]{>{\centering\arraybackslash}m{#1}}

\title{\textsc{Humanly}: A Configurable and Traceable Environment \\[4pt] for Human-AI Collaborative Writing}

\author{
\large{Shenzhe Zhu\textsuperscript{1,2,*},
Haoqian Zhang\textsuperscript{2,*},
Xu Yang\textsuperscript{1,*},
Jingyu Tang\textsuperscript{1},
Yi Nian\textsuperscript{5},} \\[4pt]
Xiaoxue Du\textsuperscript{4},
\large{Shu Yang\textsuperscript{6},
Alex Pentland\textsuperscript{3,4},
Joachim Baumann\textsuperscript{3},
Jiaxin Pei\textsuperscript{1,3,$\dag$}} \\[6pt]
\textsuperscript{1}UT Austin \quad
\textsuperscript{2}University of Toronto \quad
\textsuperscript{3}Stanford University \quad
\textsuperscript{4}MIT \quad
\textsuperscript{5}USC \quad
\textsuperscript{6}KAUST \\[4pt]
\textsuperscript{*}Equal Contribution \quad \textsuperscript{$\dag$}Corresponding Author \\[2pt]
\faDesktop~\url{https://writehumanly.net/} \quad
\faGithub~\url{https://github.com/Humanly-Lab/humanly} \\
\faEnvelope~\texttt{shenzhe@utexas.edu; jiaxinpei@utexas.edu}
}

\date{}

\begin{document}
\bibliographystyle{plainnat}

\setlength{\droptitle}{-0.6in}
\maketitle
\thispagestyle{empty}
\vspace{-2em}
\begin{abstract}
Teachers, conference chairs, and public readers all judge writing from limited evidence, seeing only a finished document and not the process that produced it.
Final text alone cannot reveal whether a document was produced through human typing, AI generation, or mixed human-AI collaboration. Existing process-tracking tools help, but many are tied to host-document histories, provide coarse activity records, and offer limited control over the writing environment. \textsc{Humanly} is a writing platform that makes the writing process itself the evidence. Users configure writing environments for personal documents or assigned tasks and draft in a workspace that records writing activity and in-platform AI assistance. \textsc{Humanly} can package a completed session into a sealed writing certificate with configuration-aware anomaly behavior review. It can support writing scenarios such as course assignments, peer review, and personal certification. Our user study shows that \textsc{Humanly} is helpful across roles, and a red-teaming study shows that the \textsc{Humanly} Typing Detector distinguishes human hand typing from automated typing.
\end{abstract}

\begin{figure}[H]
\centering
\includegraphics[width=\textwidth]{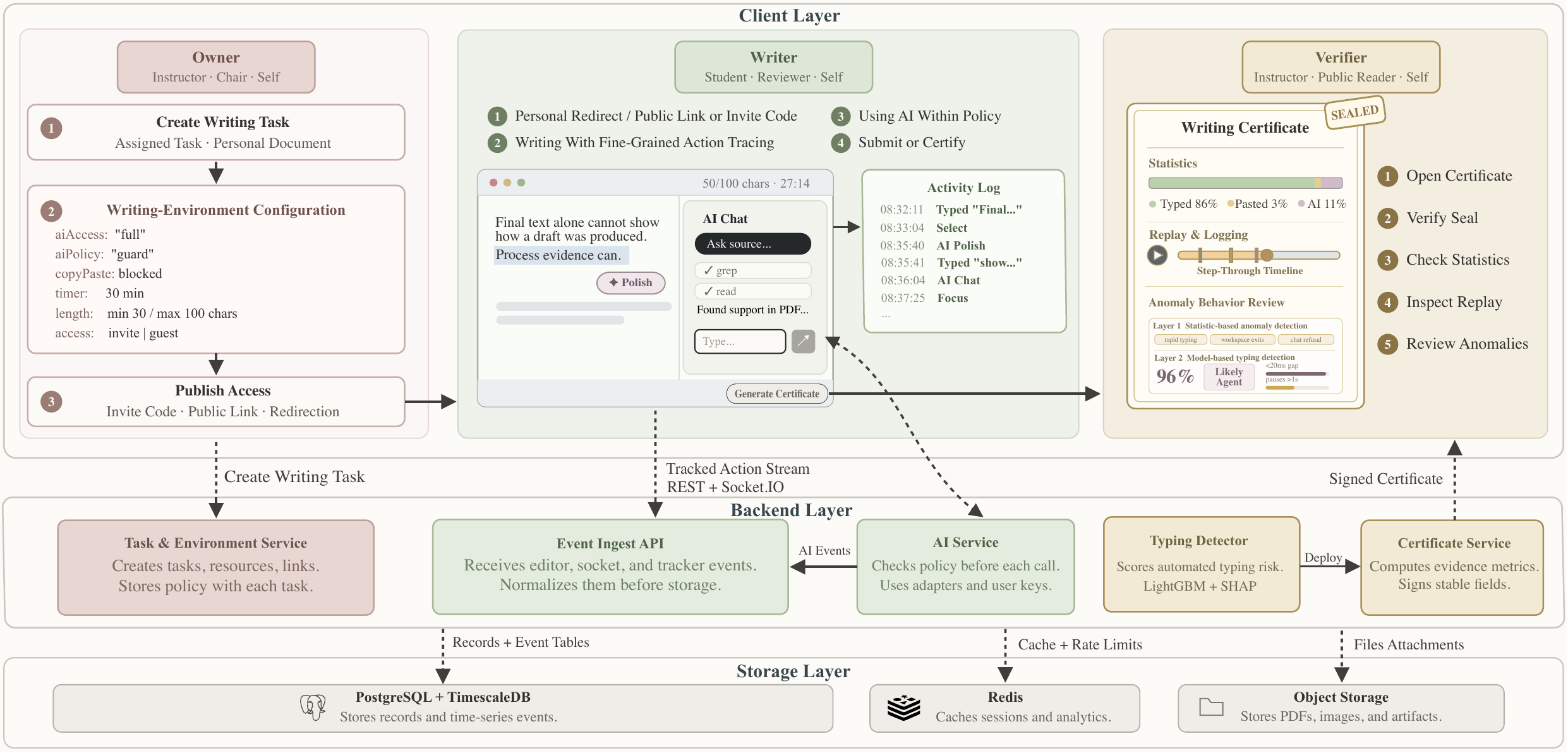}
\caption{\textsc{Humanly} overview. Owners configure a creation mode and writing policy; writers draft under permitted controls while \textsc{Humanly} records the session; the backend serves AI responses, coordinates enabled detectors for anomaly behavior review, and issues sealed certificates with logs, replay, and detector results for verifier inspection.}
\label{fig:overview}
\end{figure}

\section{Introduction}
\label{sec:intro}
Writing authenticity has become a practical problem in settings where decisions are made from final documents alone. Instructors receive student assignments as documents and need to decide whether students followed course AI-use rules; conference chairs may need to inspect whether reviewers used AI; and public readers may question whether a social media post was written by a person or AI. In each case, the final text is visible, but the process behind it is not. We define writing authenticity as whether the production history of a document matches its authorship and AI-use claims: \textit{who wrote what, with which tools, and under what rules}. This question has become harder as human--AI writing has moved from an edge case to a routine practice. Writers use AI for editing, translation, source understanding, brainstorming, and rewriting. These workflows are not equivalent to asking an AI model to produce the final text.

Post-hoc detectors classify final text without observing how it was made, leaving them blind to the writing process. They can misclassify non-native English writing as AI-generated \citep{liang2023gpt}, make inconsistent errors under translation and obfuscation \citep{weberwulff2023testing}, and lose reliability under paraphrasing or simple text manipulation \citep{sadasivan2025reliably,perkins2024simple}. Watermarking is also limited to watermark-enabled generation, not arbitrary mixed human--AI workflows \citep{kirchenbauer2023watermark}. These findings show why a detector score is too little evidence for policy decisions about mixed writing. A post-hoc detector only estimates whether final text appears AI-generated; it cannot distinguish policy-compliant polishing or translation of a human draft from substantive AI generation followed by human editing. Table~\ref{tab:final-text-motivation} summarizes four such histories with different policy implications.

\begin{table}[h]
\centering
\small
\setlength{\tabcolsep}{5pt}
\renewcommand{\arraystretch}{1.18}
\begin{tabularx}{\linewidth}{
  @{}
  >{\raggedright\arraybackslash}p{0.31\linewidth}
  >{\raggedright\arraybackslash}X
  @{}
}
\toprule
\textcolor{humanlyrosedark}{\textbf{Writing Process}} &
\textcolor{humanlyrosedark}{\textbf{Why post-hoc detection is insufficient}} \\
\midrule

\rowcolor{humanlyrow}
\textbf{Human draft + AI polish} &
Cannot reliably demonstrate how AI is involved. \\

\rowcolor{white}
\textbf{Human draft + AI translation} &
Cannot recover the human-authored source before translation. \\

\rowcolor{humanlyrow}
\textbf{Human-written AI-style text} &
Style cues can create false suspicion. \\

\rowcolor{white}
\textbf{AI draft + human polish} &
Cannot show that substantive AI generation preceded human editing. \\

\bottomrule
\end{tabularx}

\caption{Writing process that post-hoc detection cannot reliably identify.}
\label{tab:final-text-motivation}
\end{table}

Process-tracking tools provide more relevant evidence than final-text detectors because they record parts of the writing process. Revision-history replay and authorship-report products (Table~\ref{tab:feature-comparison}) can show that text appeared over time, or that some text was typed, pasted, or AI-assisted. However, important practical gaps remain. Many replay-oriented tools are tied to host-document or form environments such as Google Docs, so the process record is often coarse: typed and pasted text may be visible, but workspace activity, detailed AI assistance, and anomaly patterns are often missing or only partially represented. These host environments also limit configurability: the task owner cannot always define the writing environment itself across AI, resource, timing, and length controls. Other systems expose process tracking as a separate product beside AI detection or writing assistance rather than connected in one writing workflow.

We introduce \href{https://writehumanly.net/}{\textsc{Humanly}}, a writing platform that combines configurable writing environments, fine-grained activity logs, and certificates to make the writing process reviewable end to end (Figure~\ref{fig:overview}). \textsc{Humanly} supports three system roles. Owners create personal documents or publish tasks for others; writers draft in the workspace while \textsc{Humanly} records writing activity; and verifiers inspect sealed writing certificates as supporting evidence.
\textsc{Humanly} is designed around four goals. The first is \textbf{Flexibility}. \textsc{Humanly} should not impose one default writing policy; owners configure writing environments through 14 setting families covering AI policy, resources, budgets, constraints, anomaly behavior review, and access. The second is \textbf{Accountability}. \textsc{Humanly} makes the writing process inspectable without replacing human judgment. The activity log and authorship statistics show how the final text was composed from typed, pasted, and AI-assisted process components, while anomaly behavior review can combine enabled Anomaly Pattern signals with the \textsc{Humanly} Typing Detector for automated-typing risk. A computer-use agent (CUA), for example, can operate software on a user's behalf through graphical or command-line actions.\footnote{Examples include \href{https://developers.openai.com/codex/app/computer-use}{OpenAI Codex computer use} and \href{https://code.claude.com/docs/en/computer-use}{Claude Code computer use}.} The third is \textbf{Verifiability}. Evidence should be issued by the platform rather than asserted by the writer. \textsc{Humanly} stores session evidence and environment settings in the certificate. An Ed25519 signature protects selected certificate fields so viewers can detect external modification. The fourth is \textbf{Compatibility}. \textsc{Humanly} is built for more than one deployment pattern: the same provenance model supports assigned tasks and personal documents, invite-code and public-link distribution, signed-in and guest writing, and MIT-licensed self-deployment.

\section{System Architecture and Deployment}
\label{sec:architecture}
\textsc{Humanly} is a TypeScript monorepo whose shared schemas connect writing-environment configuration, event ingestion, AI services, certificate generation, detector orchestration, and storage (Figure~\ref{fig:overview}). We call machine-readable backend records \emph{events} and their human-readable log entries \emph{activities}.

\noindent\textbf{Runtime and Storage.}
First-party clients and the embeddable tracker connect to an Express + Socket.IO backend. Authenticated APIs serve the first-party product workflow, while an open-CORS route ingests external tracker events. PostgreSQL stores product records, TimescaleDB hypertables store high-frequency native editor and tracker events, Redis supports caching and rate limits, and resources use local or cloud storage.

\noindent\textbf{Task and Environment Service.}
The task service stores the active writing environment with each personal document or assigned task, covering AI, resources, writing constraints, detectors, and access. AI and certificate services resolve this saved record so assistance and later review use the same policy. Route guards enforce ownership, participation, and link or guest eligibility before writes.

\noindent\textbf{Event Ingest and AI Service.}
Native editor events enter \texttt{document\_events}, while external tracker events use a batched ingestion route. Before provider dispatch, the AI service resolves the saved environment, checks chat or polish availability, selects the provider/model, and applies the token budget. Responses stream through Socket.IO, while AI sessions, quick-action decisions, and policy refusals are persisted for later review.

\noindent\textbf{Certificate Service and Anomaly Detector.}
At issuance, the certificate service freezes the session boundary, computes authorship statistics and detector results, and stores the environment snapshot and verification token; logs and replay are served against the same boundary. The service reads the saved detector configuration so disabled detectors remain explicit. Anomaly Pattern derives five statistic-based signals from recorded events and policy conditions. The \textsc{Humanly} Typing Detector sends session events to a LightGBM inference service~\citep{ke2017lightgbm}, which estimates automated-typing probability from keystroke rhythm and editing behavior and returns SHAP-based contributing features~\citep{lundberg2017unified}. Insufficient timing data produces an inconclusive result. Ed25519 signs the protected certificate fields, including detector results when present, and a backend endpoint exposes the public verification key.

\noindent\textbf{Deployment.}
For local self-hosting, \textsc{Humanly} provides a one-line bootstrap command:
\texttt{curl -fsSL }\url{https://writehumanly.net/install.sh}\texttt{ \textbar{} sh}
The installer prepares Docker/Compose, source checkout, local secrets, storage, and an admin account, then starts the database, cache, backend, and portals. Production deployments replace local URLs, email, and storage settings.

\begin{figure}[htpb]
\centering
\begin{minipage}[t]{0.49\textwidth}
\centering
\includegraphics[width=\linewidth]{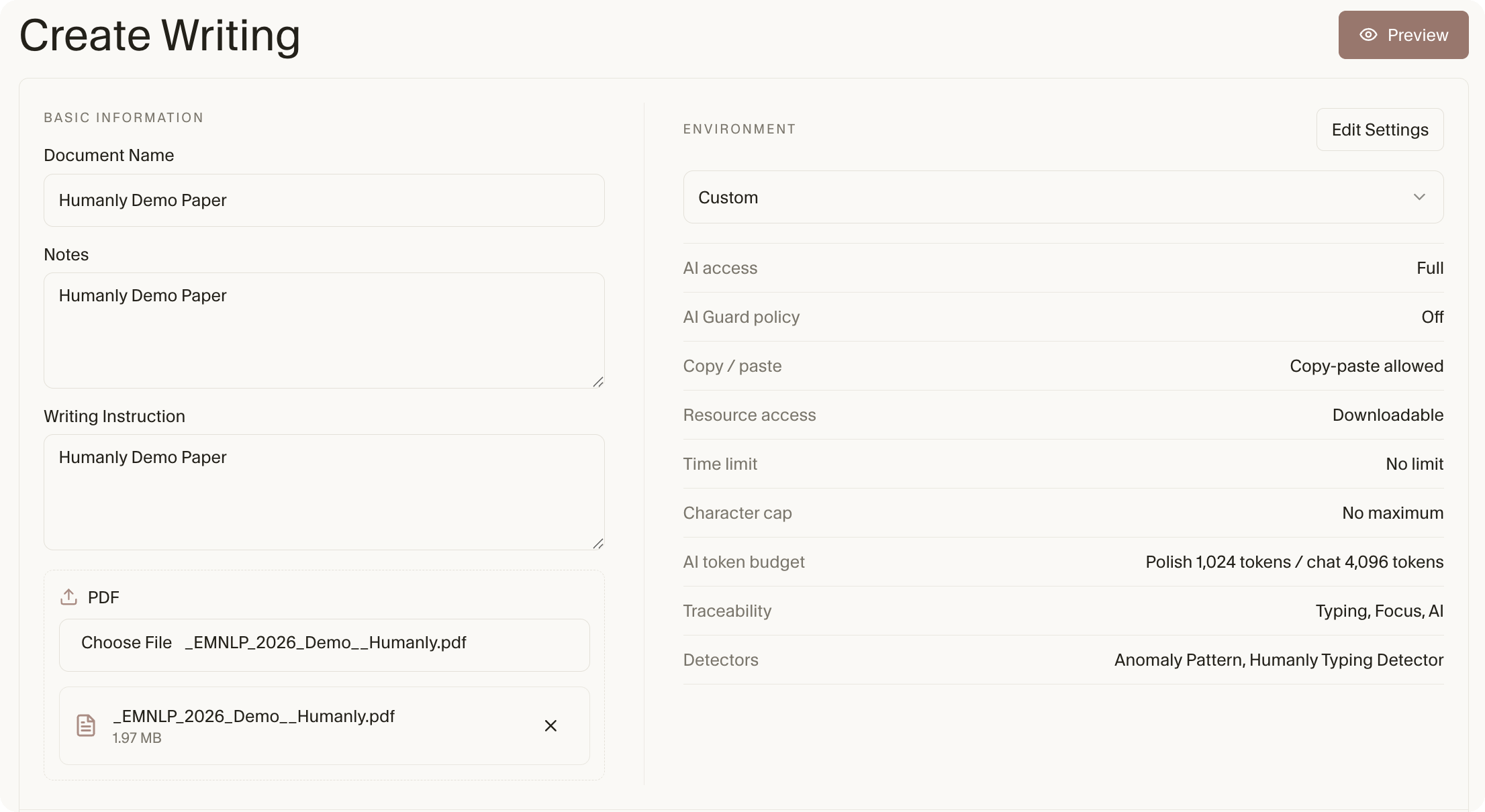}
\scriptsize \textbf{(a)} Writing environment configuration
\end{minipage}\hfill
\begin{minipage}[t]{0.49\textwidth}
\centering
\includegraphics[width=\linewidth]{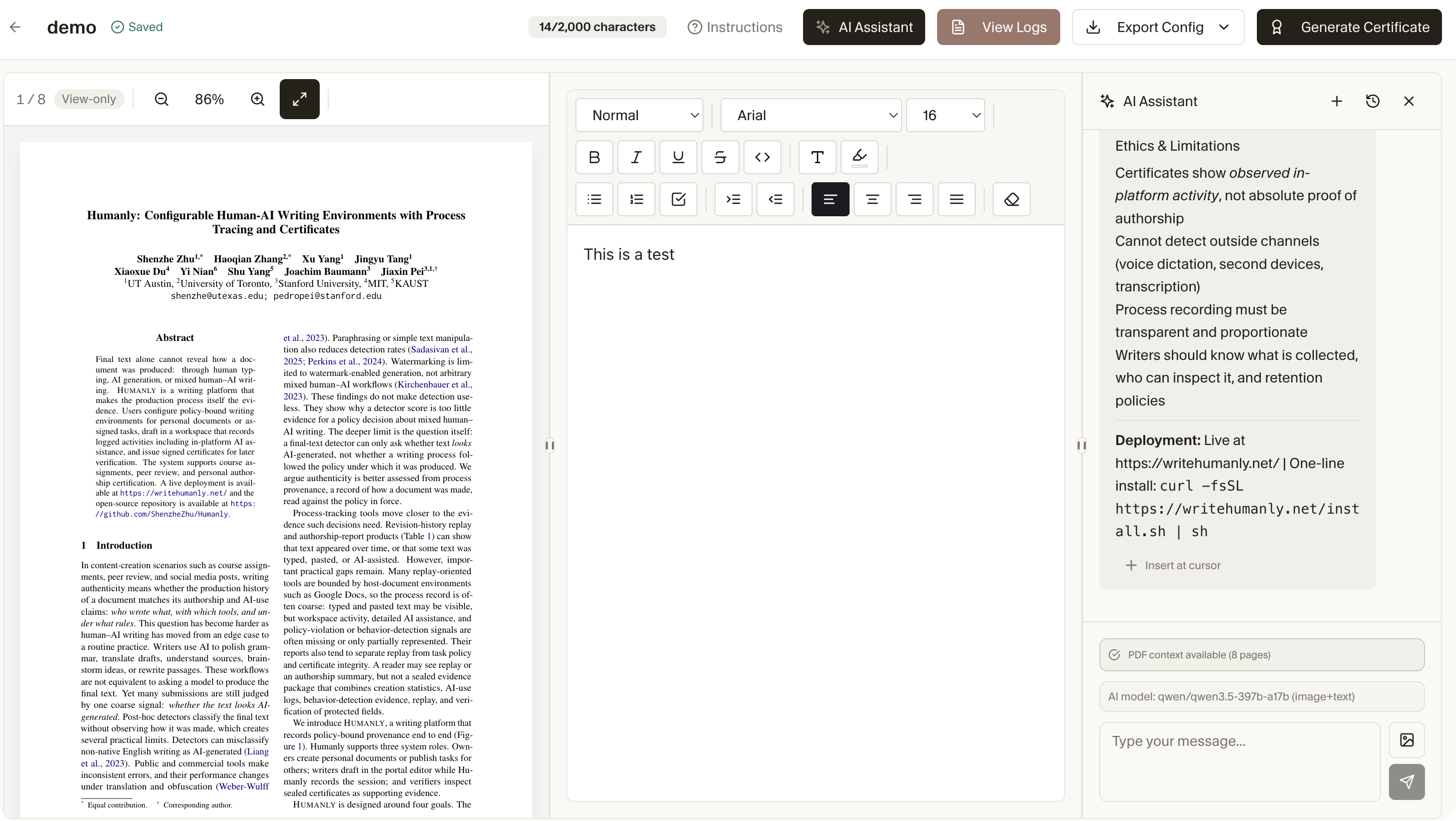}
\scriptsize \textbf{(b)} Tracked workspace
\end{minipage}

\vspace{0.6em}

\begin{minipage}[t]{0.49\textwidth}
\centering
\includegraphics[
  width=\linewidth,
  trim=0 2.5cm 0 0,
  clip
]{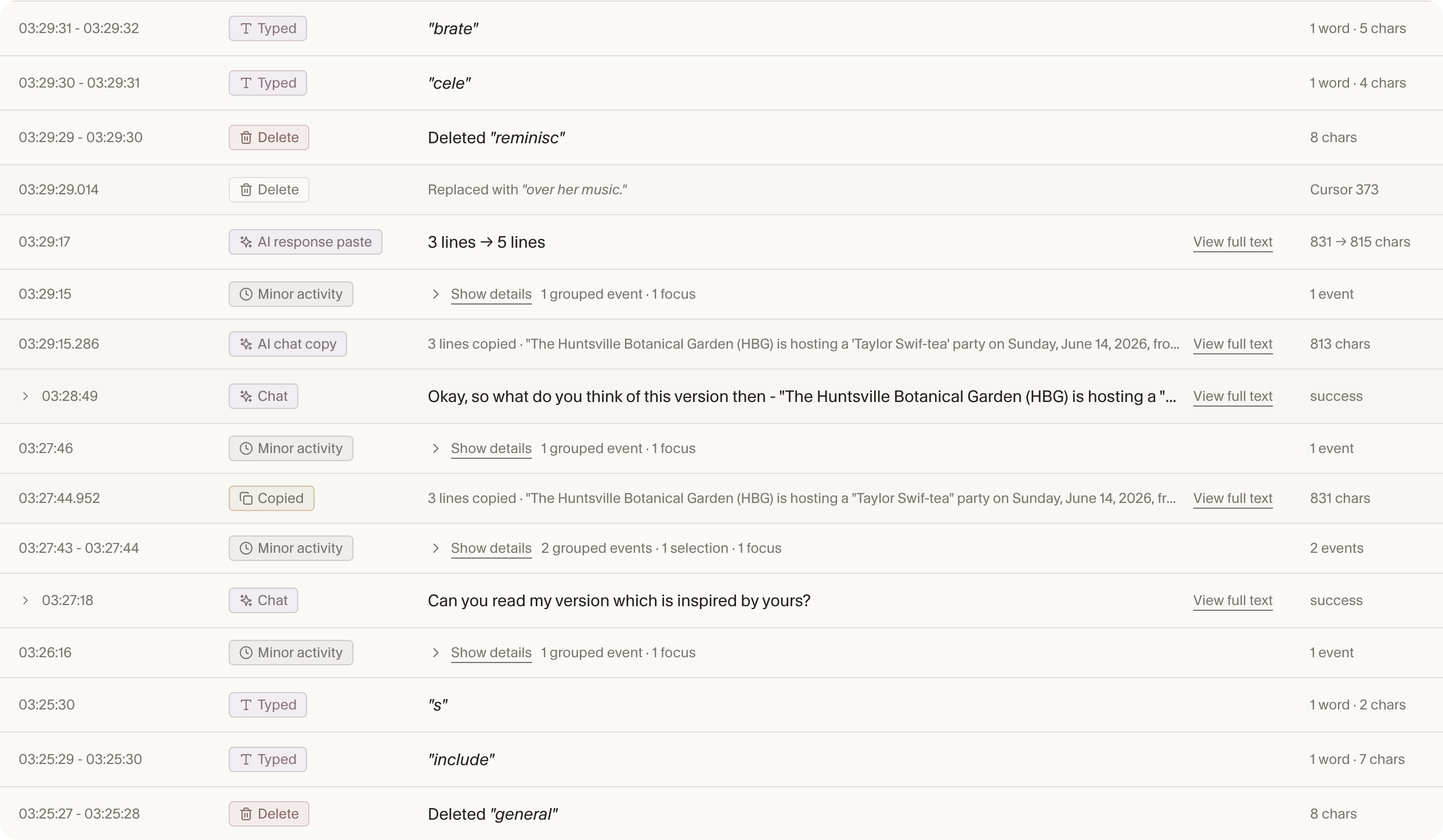}
\scriptsize \textbf{(c)} Activity log
\end{minipage}\hfill
\begin{minipage}[t]{0.49\textwidth}
\centering
\includegraphics[width=\linewidth,
  trim=0 0 0 0,
  clip]{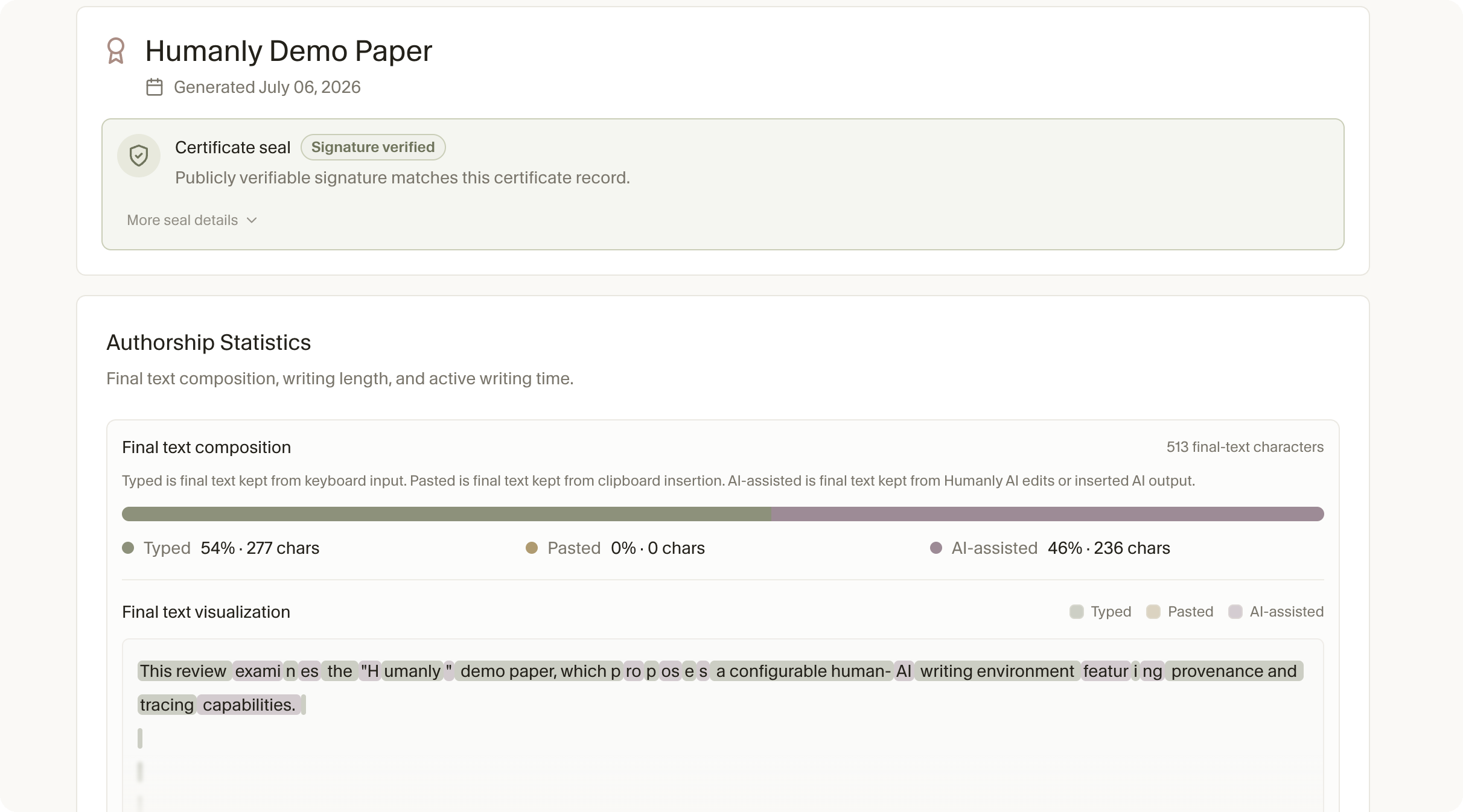}
\scriptsize \textbf{(d)} Certificate
\end{minipage}
\caption{\textsc{Humanly} interfaces: (a) writing environment configuration; (b) tracked workspace in full mode with PDF resources, editor, and AI assistant; (c) activity log; and (d) certificate with authorship statistics and signature.}
\label{fig:workflow-screens}
\end{figure}

\begin{table}[t]
\centering
\small
\setlength{\tabcolsep}{3pt}
\renewcommand{\arraystretch}{0.95}
\resizebox{\textwidth}{!}{%
\begin{tabular}{@{}llcccccccc@{}}
\toprule
Area & Feature
& \textbf{\textsc{Humanly}}
& \href{https://www.turnitin.ca/products/feedback-studio/clarity}{Turnitin}
& \href{https://support.grammarly.com/hc/en-us/articles/29548735595405-Introducing-Authorship}{Grammarly}
& \href{https://support.gptzero.me/articles/7001890416-what-is-the-google-docs-writing-report-for-origin}{GPTZero}
& \href{https://draftback.com/}{Draftback}
& \href{https://www.briskteaching.com/inspect-writing}{Brisk}
& \href{https://integrito.ai/}{Integrito}
& \href{https://papertrailacademic.com/inspect/}{PaperTrail} \\
\midrule
\multirow{4}{*}{\begin{tabular}[c]{@{}c@{}}Writing\\Environment\end{tabular}}
& Configurable rules & \yesfeature & \yesfeature & \nofeature & \nofeature & \nofeature & \nofeature & \nofeature & \nofeature \\
& Task assignment & \yesfeature & \yesfeature & \nofeature & \nofeature & \nofeature & \nofeature & \nofeature & \nofeature \\
& AI policy & \yesfeature & \yesfeature & \nofeature & \nofeature & \nofeature & \nofeature & \nofeature & \nofeature \\
& Open source & \yesfeature & \nofeature & \nofeature & \nofeature & \nofeature & \nofeature & \nofeature & \nofeature \\
\midrule
\multirow{5}{*}{Activity Log}
& Text editing & \yesfeature & \yesfeature & \yesfeature & \yesfeature & \yesfeature & \yesfeature & \yesfeature & \yesfeature \\
& Clipboard & \yesfeature & \yesfeature & \yesfeature & \yesfeature & \nofeature & \yesfeature & \yesfeature & \yesfeature \\
& Workspace state & \yesfeature & \nofeature & \nofeature & \nofeature & \nofeature & \nofeature & \nofeature & \nofeature \\
& AI assistance & \yesfeature & \yesfeature & \yesfeature & \nofeature & \nofeature & \nofeature & \nofeature & \nofeature \\
& Anomaly pattern & \yesfeature & \yesfeature & \nofeature & \yesfeature & \nofeature & \nofeature & \yesfeature & \yesfeature \\
\midrule
\multirow{4}{*}{Certificate}
& Replay & \yesfeature & \yesfeature & \yesfeature & \yesfeature & \yesfeature & \yesfeature & \yesfeature & \yesfeature \\
& Authorship statistics & \yesfeature & \yesfeature & \yesfeature & \yesfeature & \yesfeature & \yesfeature & \yesfeature & \yesfeature \\
& Typing detector & \yesfeature & \nofeature & \nofeature & \nofeature & \nofeature & \nofeature & \nofeature & \nofeature \\
& Integrity seal & \yesfeature & \nofeature & \nofeature & \nofeature & \nofeature & \nofeature & \nofeature & \nofeature \\
\bottomrule
\end{tabular}}

\caption{Feature comparison of writing-authenticity systems by writing environment, activity log, and certificate. Competitor columns refer to the linked product documentation; \(\checkmark\) means supported, and blank cells mean not found in public documentation. Configurable rules = owner-set environment constraints; AI policy = visible AI-use modes and guard rules; workspace state = focus, selection, or workspace leave/return events; anomaly pattern = statistic-based anomaly behavior pattern derived from logged events and active policy; typing detector = estimate of whether typed input resembles human hand typing or automated typing; integrity seal = Ed25519 certificate signature that can be checked with \textsc{Humanly}'s public key.}
\label{tab:feature-comparison}

\end{table}

\section{Workflow and Use Cases}

\subsection{Workflow Design}
\label{subsec:workflow}
Figure~\ref{fig:overview} organizes the workflow around three roles: Owner, Writer, and Verifier. These are not fixed account types: assigned tasks may distribute them across users, while personal documents may combine the Owner and Writer roles. Figure~\ref{fig:workflow-screens} follows the corresponding interfaces from setup through writing and certificate review.

\noindent\textbf{Owner.}
The Owner role begins when a user creates a writing environment. Personal-document owners configure a private document in the \href{https://app.writehumanly.net/}{Writer Portal}; assigned-task owners configure a shared task in the \href{https://admin.writehumanly.net/}{Publisher Portal} and distribute it through invite codes or public links. Across both paths, owners configure 14 setting families, including resources, constraints, access, four AI modes (off, polish-only, chat-only, and full), and two detectors for anomaly behavior review: Anomaly Pattern and \textsc{Humanly} Typing Detector. Table~\ref{tab:configuration-settings} in Appendix~\ref{app:config-capture} summarizes the setup options. \textsc{Humanly} saves this configuration with the session, so the certificate can be interpreted against the rules active during writing.

\noindent\textbf{Writer.}
The Writer role begins when a user enters a configured workspace, either through their own personal document or through an assigned task shared by another owner. Before drafting, the writer is shown the task instructions and writing rules defined by the owner-set policy.
With full mode and source resources enabled, the workspace presents three panels: source PDFs or resources, the editor, and the AI assistant. Writers can use chat to ask questions about the provided resources and apply four AI quick actions to selected text during drafting. The activity log exposes 27 reviewable activity labels across text editing, clipboard activity, workspace status, AI assistance, and anomaly patterns. Table~\ref{tab:process-capture} in Appendix~\ref{app:config-capture} counts these labels by category. When writing is complete, the writer submits the task or generates a certificate from the recorded session.

\noindent\textbf{Verifier.}
Certificate review begins after issuance. For assigned tasks, issuance follows submission; for personal documents, the writer generates the certificate after finishing the document. A verifier may be the original writer, the task owner, or any public reader with a shared verification link. The certificate evidence includes authorship statistics, environment settings, an activity log, replay, the anomaly behavior review configured for that writing environment, and an Ed25519 integrity seal. Authorship statistics summarize final-text composition and process input volume; environment settings preserve the policy active during writing; the activity log presents recorded activities; and replay shows how the document changed over time. When both detectors are enabled, anomaly behavior review includes Anomaly Pattern signals and the \textsc{Humanly} Typing Detector score; disabled detectors are shown as not enabled. The integrity seal signs selected certificate fields with Ed25519, allowing verifiers to use \textsc{Humanly}'s public key to check whether those fields were modified after issuance.

\subsection{Use Cases}
\label{subsec:usecases}
\noindent\textbf{Course assignments.}
Course assignments follow a familiar learning-management workflow: an instructor publishes a task, students complete it, and the instructor reviews the submission. Standard assignment systems collect the final file, but they rarely show whether students followed AI-use rules while writing. \textsc{Humanly} preserves the same assignment flow while adding a configurable writing environment and process evidence that can surface anomaly patterns and automated typing behavior when possible rule violations need inspection.

\noindent\textbf{Peer review.}
Peer review has a similar publish-write-review flow: an area chair assigns a paper, reviewers write reviews, and chairs inspect the results. AI-assisted reviewing is becoming part of this workflow at several academic conferences~\citep{baumann2026stop}.\footnote{\href{https://neurips.cc/Conferences/2026/ai-reviewing-experiment}{NeurIPS 2026 AI-assisted reviewing experiment}; \href{https://icml.cc/Conferences/2026/LLM-Policy}{ICML 2026 LLM reviewing policy}.} Even under such policies, compliance often depends on reviewers following the rules without direct process evidence~\citep{saha2026policies}. \textsc{Humanly} is useful here because chairs can provide an in-platform AI assistant, record how it was used, and inspect process evidence when review quality is questioned.

\noindent\textbf{Social media posting.}
Social media posts are often judged from final text alone, and recent reporting describes human writers accused of AI use when their prose appears polished or machine-like.\footnote{\href{https://whyy.org/segments/how-not-to-be-mistaken-for-a-chatbot/}{WHYY, ``How Not to Be Mistaken for a Chatbot''}; \href{https://nymag.com/intelligencer/article/the-people-getting-falsely-accused-of-using-ai-to-write.html}{New York Magazine, ``The People Falsely Accused of Using AI''}.} In this setting, \textsc{Humanly} is useful because a poster can draft as a personal document and share a signed certificate link with readers. Instead of inferring authorship from style alone, readers can inspect the observed writing process behind the post.

\section{Comparison with Existing Systems}
\label{sec:comparison}
Table~\ref{tab:feature-comparison} compares \textsc{Humanly} with seven related writing-authenticity systems that provide process evidence, authorship reports, or replay for written documents. The rows follow three product areas: a configurable writing environment, a native activity log, and a certificate layer. Ratings indicate whether each feature is supported in public product documentation.
Overall, the closest comparator is Turnitin Clarity: it supports instructor-managed assignments in which students write under AI-use guidance and instructors receive writing-process reports. Grammarly Authorship adds authorship tracking, source attribution, shareable reports, and replay across supported writing surfaces. The other five products rely primarily on Google Docs or browser-extension workflows for writing reports and replay. We therefore score whether each system allows a task owner to configure writing policy before writing, records human and AI activity at sufficient granularity, and packages the result as certificate-style evidence.

\begin{figure}[h]
\centering
\includegraphics[width=0.94\textwidth]{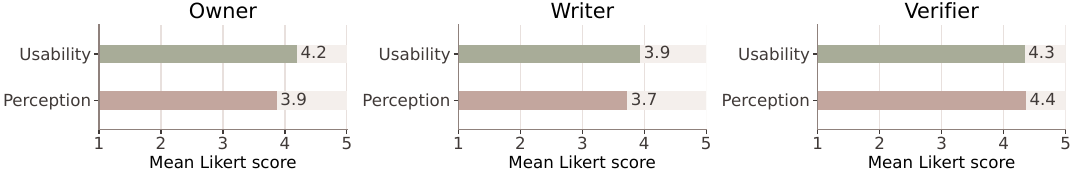}
\caption{Role-based user study results. Bars report respondent-level mean Likert scores for Usability and Perception. Scale: 1 = Strongly disagree, 2 = Disagree, 3 = Neither agree nor disagree, 4 = Agree, and 5 = Strongly agree.}
\label{fig:user-study-bars}
\end{figure}

\noindent\textbf{Writing Environment.}
\textsc{Humanly}'s first difference is that it offers a configurable writing environment rather than depending on a host editor's inherited policies. Owners set resource, timing, length, access, and AI rules before writing. Turnitin Clarity is strongest in this category because it supports instructor-controlled assignment settings and AI-use guidance for student writing. \textsc{Humanly} is released for inspection, adaptation, and self-deployment.

\noindent\textbf{Activity Log.}
\textsc{Humanly}'s second difference is the scope of the fine-grained activity log. The workspace logs text editing, clipboard, workspace state, AI assistance, and anomaly patterns. Most systems in the table provide text-editing and clipboard records, but workspace state is unique to \textsc{Humanly}: it records how the writer uses the workspace, including focus, blur, workspace leave, and return events. Four competitors expose narrower anomaly cues, such as playback flags, writing-pattern evaluation, suspicious-event summaries, or struggle moments.

\noindent\textbf{Certificate.}
In the certificate layer, \textsc{Humanly} links authorship statistics, writing replay, anomaly behavior review, task policy, and the integrity seal. \textsc{Humanly} differs in that these fields are bundled into one certificate package rather than split across separate reports or tools. GPTZero illustrates this fragmentation: its Chrome extension offers Google Docs writing logs, an AI assistant, and AI scan, but its AI-composition verification still relies on a prediction scan over final text rather than attribution from the recorded process and in-extension assistant actions.\footnote{In an illustrative manual check of the GPTZero Google Docs workflow (June 2026), text generated through GPTZero's in-extension AI assistant was still rated as human by the separate scan feature.} Model-based human typing detection and the integrity seal are unique to \textsc{Humanly} in this comparison. The other products do not address whether typed input was produced by human hand typing or automated workspace operation. They also do not protect certificate evidence against later tampering with a public-key signature.

\section{Evaluation}
\label{sec:evaluation}
We evaluate \textsc{Humanly} through a role-based user study and a red-teaming study. The user study tests whether the system is usable across the Owner, Writer, and Verifier roles; the red-teaming study tests whether the \textsc{Humanly} Typing Detector separates human hand typing from automated workspace operation in a controlled writing task.

\subsection{Role-Based User Study}
\label{sec:user-study}
The role-based user study examines the three roles in Figure~\ref{fig:overview}: Owner, Writer, and Verifier. We recruit three independent groups of Prolific AI Taskers in the Structured Writing category\footnote{\href{https://researcher-help.prolific.com/en/articles/445229-participants-skilled-at-ai-tasks}{Prolific AI Taskers, Structured Writing}}. Each group receives one survey, and quality review leaves 30 valid responses per role. Survey items use a five-point agreement scale from Strongly disagree to Strongly agree. Across roles, \textbf{Usability} measures whether participants can use or interpret the relevant interface. \textbf{Perception} captures role-specific judgments: Owners rate policy transparency and confidence in rule compliance; Writers rate accountability, authorship proof, and recording anxiety; Verifiers rate fairness and trust in process evidence. Detailed setup appears in Appendix~\ref{app:user-study-surveys}.

\noindent\textbf{Analysis.}
Figure~\ref{fig:user-study-bars} shows positive Usability and Perception ratings across all three roles, with every mean above 3 on the five-point agreement scale. Verifiers give the strongest ratings (\(M=4.34\) for Usability and \(M=4.36\) for Perception). This suggests that readers who did not produce the writing can still understand the evidence package. Writer Perception is lower (\(M=3.73\)) but still positive. This result reflects the tradeoff in the writer role: participants valued clearer rules and evidence they could later share, but some were also uneasy about detailed recording while they wrote.

\subsection{Red-Teaming Study}
\label{sec:redteam}
Motivated by recent work on the security of computer-use agents (CUAs)~\citep{chen2025survey,liao2025redteamcua}, we perform a red-teaming study to evaluate the \textsc{Humanly} Typing Detector.
We test one configurable detector in \textsc{Humanly}'s anomaly behavior review under a policy that requires writers to operate the workspace directly; delegating browser control to a CUA is therefore non-compliant. The detector estimates whether typed interaction resembles human hand typing or automated typing.

\noindent\textbf{Data Collection.}
We collect human-operated submissions through Prolific. For agent-operated submissions, we use OpenAI Codex's computer-use capability with GPT-5.5 as the base model to control the workspace. Copy-paste remains allowed, and a 20-minute limit standardizes task completion. Detailed settings appear in Appendix~\ref{app:redteam-details}.

\noindent\textbf{Baselines and Metrics.}
We compare the \textsc{Humanly} Typing Detector against three zero-shot LLM-as-judge baselines that score serialized activity logs (Table~\ref{tab:detector-eval}). Human FPR is our primary operating constraint: falsely flagging a compliant writer can create unwarranted suspicion, so the detector must keep this error low while identifying automated typing. We therefore use a conservative threshold and report human false-positive rate (FPR; lower is better), agent true-positive rate (TPR; higher is better), AUROC (higher is better), and inference time (lower is better). More details are in Appendix~\ref{app:redteam-details}.

\begin{table}[h]
\centering
\small
\setlength{\tabcolsep}{14pt}
\begin{tabular}{@{}lcccc@{}}
\toprule
\textcolor{humanlyrosedark}{\textbf{Method}}
& \textcolor{humanlyrosedark}{\textbf{Human FPR \(\downarrow\)}}
& \textcolor{humanlyrosedark}{\textbf{Agent TPR \(\uparrow\)}}
& \textcolor{humanlyrosedark}{\textbf{AUROC \(\uparrow\)}}
& \textcolor{humanlyrosedark}{\textbf{Inference time (s) \(\downarrow\)}} \\
\midrule
LLM-as-Judge (GPT-5.1) & \textbf{0.0}\% & 15.0\% & 0.563 & 3.4 \\
LLM-as-Judge (Claude Sonnet 4.6) & \textbf{0.0}\% & 60.0\% & 0.838 & 12.7 \\
LLM-as-Judge (Claude Opus 4.8) & 5.9\% & \textbf{90.0}\% & 0.965 & 11.5 \\
\rowcolor{humanlyrow}
\textbf{Ours (LightGBM)} & \textbf{0.0}\% & 85.0\% & \textbf{0.994} & \textbf{0.0024} \\
\bottomrule
\end{tabular}
\arrayrulecolor{black}
\caption{\textsc{Humanly} Typing Detector compared with LLM-as-judge baselines on red-team submissions.}
\label{tab:detector-eval}
\end{table}

\noindent\textbf{Results.}
At this operating point, \textsc{Humanly} achieves 0.0\% human FPR, 85.0\% agent TPR, and the best AUROC. Claude Opus 4.8 reaches a higher agent TPR (90.0\%) but flags 5.9\% of human-operated submissions and has lower AUROC. \textsc{Humanly} averages 0.0024 seconds per submission, approximately \(4{,}800\times\) faster than Opus 4.8 and at least \(1{,}400\times\) faster than every LLM judge in our setup. Local LightGBM inference also avoids a paid model call for each full activity log. The detector stores SHAP-based contributing features, giving verifiers feature-level explanations of which typing behaviors drove the score.

\section{Conclusion}
\label{sec:conclusion}
\textsc{Humanly} reframes writing authenticity as policy-bound process evidence. It combines configurable writing environments, activity logs, anomaly behavior review, replay, and public-key verification. Across course assignments, peer review, and personal certification, the workflow ties owner-defined rules to evidence. Our studies show cross-role usability and reliable separation of human and automated typing with few false alarms.

\section*{Ethics and Broader Impact Statement}
\label{sec:limitations}

\noindent\textbf{Broader Impact.}
\textsc{Humanly} shifts writing-authenticity decisions away from final-text suspicion and toward process evidence. This can benefit writers whose polished, translated, or non-native English work is questioned, and it can help instructors, editors, and public readers review mixed human--AI writing with more context. The same capability creates risk: process evidence can become intrusive if recording becomes routine surveillance or certificates become disciplinary shortcuts. \textsc{Humanly} should therefore be deployed as a policy-transparent review aid, not as general monitoring or an automatic misconduct judge.

\noindent\textbf{Limitation: What Certificates Can and Cannot Show.}
A \textsc{Humanly} certificate records observed in-platform activity; the certificate is not an authorship verdict. It reports the recorded process, policy context, statistic-based anomaly detection, and, when enabled, a model-based human typing score. The certificate cannot prove intention or rule out every outside channel. Page-visibility records show that the \textsc{Humanly} workspace was hidden or visible again; they do not identify which tab, website, application, or device the writer used while away. A writer can transcribe external AI output, dictate it by voice, or route text through another device. Certificates therefore support inspection of observed activity rather than proof that no outside assistance occurred. The red-teaming study evaluates browser or GUI-agent operation of the \textsc{Humanly} editor, not every off-platform assistance channel. \textsc{Humanly} only records activity inside \textsc{Humanly}; tracker-based documents carry less detail than native-editor documents, and writing outside \textsc{Humanly} leaves no record.

\noindent\textbf{Recording Governance.}
Process recording captures typing behavior, clipboard activity, workspace activity, formatting, and AI assistance. Writers should know which activity categories are logged, who can inspect them, how long they are retained, and whether public verification exposes a summary or a full replay. \textsc{Humanly} surfaces the rules at document setup and keeps only the permitted controls visible while writing. Deployments still need clear local policies for consent, access, retention, appeals, and reader training, especially in classrooms and peer review. Our studies follow the same principle: recording is disclosed up front, and participants are asked to interpret evidence without treating the certificate as an automatic verdict. Personally identifiable information is limited to participant payment and study administration.

\bibliography{custom}

\newpage
\appendix

\section{User-Study Survey Administration}
\label{app:user-study-surveys}
We administer three role-specific surveys in Deliberation.io
\citep{pei2025deliberationio,zhu2025can}. Prolific\footnote{\href{https://www.prolific.com/}{Prolific}}
routes qualified Structured Writing AI Taskers to the corresponding survey, and
participants receive a completion code upon submission. Owners create and publish a task,
Writers complete an assigned task in \textsc{Humanly}, and Verifiers review fixed
certificate screenshots and context. All agreement items use a five-point scale,
and we retain 30 valid responses per role. For Owners and Writers, we manually
inspect submitted task or certificate links for on-task completion. Because
Verifiers review fixed materials, we instead exclude submissions completed in
under five minutes (expected: 10--20 minutes). Dimension scores are respondent-level
means. ``I did not use it'' responses to optional Writer AI-usefulness items are
excluded item-wise, and the Writer anxiety item is reverse-coded before computing
Perception. The full questionnaire is available online.\footnote{\href{https://github.com/Humanly-Lab/humanly/blob/main/paper-artifacts/user-study-survey-questions.md}{\textsc{Humanly} user-study question list}.}

\section{Red-Teaming Study Details}
\label{app:redteam-details}

\noindent\textbf{Task and Agent Setup.}
We collect 20 human-operated and 20 agent-operated submissions. After quality filtering, 17 human-operated and 20 agent-operated submissions remain usable for the common evaluation. Red-team sessions use a standardized TOEFL ``Writing for an Academic Discussion'' prompt in the native \textsc{Humanly} editor, with the AI mode set to full, copy-paste allowed, a 20-minute window, and a 1000--2000 character requirement. The test prompt uses a topic disjoint from the topics used to train the \textsc{Humanly} Typing Detector. For automated-operation sessions, a computer-use agent (CUA) controls the browser and completes the \textsc{Humanly} writing task. We instantiate this condition through the OpenAI Codex platform,\footnote{\href{https://developers.openai.com/codex/app/computer-use}{OpenAI Codex computer use}.} using GPT-5.5 as the base model, under two prompting strategies: Direct and Human-like. The prompt text is maintained online.\footnote{\href{https://github.com/Humanly-Lab/humanly/blob/main/paper-artifacts/red-team-codex-prompts.md}{\textsc{Humanly} red-team Codex prompts}.}

\noindent\textbf{Baselines and Metrics.}
The LLM-as-judge baselines score each serialized activity log zero-shot, using GPT-5.1, Claude Sonnet 4.6, and Claude Opus 4.8; Claude Opus 4.8 runs with high reasoning effort. Each judge receives a generic task description with no hand-engineered detection cues or labeled examples. The judge prompt is maintained online.\footnote{\href{https://github.com/Humanly-Lab/humanly/blob/main/paper-artifacts/llm-as-judge-prompt.md}{\textsc{Humanly} LLM-as-judge prompt}.} The \textsc{Humanly} Typing Detector threshold is chosen on held-out validation submissions to keep the human false-positive rate low, since a false flag directs review toward a compliant writer. Treating agent-operated submissions as positive, Table~\ref{tab:redteam-confusion} defines the thresholded outcomes. We report four metrics. \textbf{Human FPR} is \(FP/(FP+TN)\), the fraction of human-operated submissions wrongly flagged as agent-operated. \textbf{Agent TPR} is \(TP/(TP+FN)\), the fraction of agent-operated submissions correctly flagged. \textbf{AUROC} summarizes threshold-free separation: the probability that a randomly chosen agent-operated submission receives a higher detector score than a randomly chosen human-operated one, where \(0.5\) is chance and \(1.0\) is perfect separation. \textbf{Inference time} is the mean wall-clock time to score one submission. For the LLM judges this is the per-submission API round-trip through OpenRouter (network plus model generation), whereas for the \textsc{Humanly} Typing Detector it is local feature extraction plus a single LightGBM prediction with no network call.

\begin{center}
\small
\setlength{\tabcolsep}{6pt}
\renewcommand{\arraystretch}{0.95}
\begin{tabular}{@{}lcc@{}}
\toprule
 & Pred. agent & Pred. human \\
\midrule
Actual agent & TP & FN \\
Actual human & FP & TN \\
\bottomrule
\end{tabular}
\captionof{table}{Confusion matrix for thresholded human typing detection.}
\label{tab:redteam-confusion}
\end{center}

\section{Configuration and Capture Details}
\label{app:config-capture}

Table~\ref{tab:configuration-settings} lists the main environment controls exposed by \textsc{Humanly}, and Table~\ref{tab:process-capture} counts the 27 reviewable activity labels in the native writing editor. The main paper summarizes these fields because the exact configuration differs between personal writing and assigned tasks.

\begin{table}[h]
\small
\centering
\setlength{\tabcolsep}{4pt}
\renewcommand{\arraystretch}{0.95}
\begin{tabular}{@{}M{0.27\linewidth}N{0.10\linewidth}N{0.13\linewidth}M{0.44\linewidth}@{}}
\toprule
Setting & Personal & Task & Values \\
\midrule
Files/resources & \checkmark & \checkmark & Source PDFs, task PDFs, and writer-provided resources. \\
\midrule
Resource access & \checkmark & \checkmark & Downloadable or view-only PDFs; view-only uses short-lived tokens and canvas rendering. \\
\midrule
Preset/import & \checkmark & \checkmark & Default, custom, and JSON import/export configurations. \\
\midrule
AI access & \checkmark & \checkmark & Off, polish-only, chat-only, or full assistance. \\
\midrule
AI guard policy & \checkmark & \checkmark & Optional rejection rules for agent chat in chat-only or full modes. \\
\midrule
AI provider/model & \checkmark & \checkmark & User API key; Together, OpenRouter, OpenAI, Anthropic; curated model/capability allowlists. \\
\midrule
AI budgets & Tokens & \shortstack{Tokens\\+requests} & Shortcut/chat token budgets; per-user task request cap. \\
\midrule
Copy/paste & \checkmark & \checkmark & Copy, cut, and paste allowed or blocked. \\
\midrule
Writing timer & \checkmark & \checkmark & Optional countdown; timed task attempts can auto-submit at expiry. \\
\midrule
Anomaly behavior review & \checkmark & \checkmark & Enable Anomaly Pattern, \textsc{Humanly} Typing Detector, or both. \\
\midrule
Length bounds & Max & Min/max & Final character limits for personal writing or assigned tasks. \\
\midrule
Task attempts & No & \checkmark & Single durable attempt or restart-allowed mode with a maximum attempt count. \\
\midrule
Availability & No & \checkmark & Task start and end window. \\
\midrule
Guest link & No & \checkmark & Public link can require sign-in or allow guest submissions. \\
\bottomrule
\end{tabular}
\caption{Configurable writing-environment settings. ``Task'' refers to admin-created assigned tasks and public share links.}
\label{tab:configuration-settings}
\end{table}

\begin{table}[h]
\small
\centering
\setlength{\tabcolsep}{4pt}
\renewcommand{\arraystretch}{0.95}
\begin{tabular}{@{}M{0.26\linewidth}N{0.10\linewidth}M{0.58\linewidth}@{}}
\toprule
Category & Count & Native editor \\
\midrule
Text editing & 5 & Typed text; deletions; replacements; line breaks; blank-line insertions. \\
\midrule
Clipboard / formatting & 4 & Paste; copy; cut; formatting changes. \\
\midrule
Workspace status & 5 & Workspace leave; workspace return; editor focus; editor unfocus; text selections. \\
\midrule
AI assistance & 8 & AI chat; AI quick actions (Fix grammar; Improve writing; Simplify; Make formal); AI inserted; AI chat copy; AI response paste. \\
\midrule
Anomaly patterns & 5 & Rapid text accumulation; untracked text source; frequent workspace exits; blocked copy-paste attempts; chat refusals. \\
\bottomrule
\end{tabular}
\caption{Reviewable activity labels in the native writing editor and certificate evidence. The five anomaly patterns are derived from logged events and certificate metrics rather than additional raw event types.}
\label{tab:process-capture}
\end{table}

\end{document}